# Machine Learning-driven Analysis of Gastrointestinal Symptoms in Post-COVID-19 Patients

Maitham G. Yousif*[1]  ,Fadhil G. Al-Amran[2], Salman Rawaf[3] , Mohammad Abdulla Grmt[4]

[1]Biology Department, College of Science, University of Al-Qadisiyah, Iraq, Visiting Professor in Liverpool John Moors University, Liverpool, United Kingdom

[2]Cardiovascular Department, College of Medicine, Kufa University, Iraq

[3]Professor of Public Health Director, WHO Collaboration Center, Imperial College, London, United Kingdom

[4]Al-Sadder Teaching Hospital, Al-Najaf Health office, Najaf, Iraq

**Abstract**

The COVID-19 pandemic, caused by the novel coronavirus SARS-CoV-2, has posed significant health challenges worldwide. While respiratory symptoms have been the primary focus, emerging evidence has highlighted the impact of COVID-19 on various organ systems, including the gastrointestinal (GI) tract. This study, based on data from 913 post-COVID-19 patients in Iraq collected during 2022 and 2023, investigates the prevalence and patterns of GI symptoms in individuals recovering from COVID-19 and leverages machine learning algorithms to identify predictive factors for these symptoms. The research findings reveal that a notable percentage of post-COVID-19 patients experience GI symptoms during their recovery phase. Diarrhea emerged as the most frequently reported symptom, followed by abdominal pain and nausea. Machine learning analysis uncovered significant predictive factors for GI symptoms, including age, gender, disease severity, comorbidities, and the duration of COVID-19 illness. These findings underscore the importance of monitoring and addressing GI symptoms in post-COVID-19 care, with machine learning offering valuable tools for early identification and personalized intervention. This study contributes to the understanding of the long-term consequences of COVID-19 on GI health and emphasizes the potential benefits of utilizing machine learning-driven analysis in predicting and managing these symptoms. Further research is warranted to delve into the mechanisms underlying GI symptoms in COVID-19 survivors and to develop targeted interventions for symptom management.

**Keywords**: COVID-19, gastrointestinal symptoms, machine learning, predictive factors, post-COVID-19 care, long COVID.

*Corresponding author: Maithm Ghaly Yousif  matham.yousif@qu.edu.iq   m.g.alamran@ljmu.ac.uk

**Introduction**

The COVID-19 pandemic, caused by the novel coronavirus SARS-CoV-2, has brought about a multitude of health challenges and continues to be a subject of extensive research worldwide. Beyond its immediate respiratory manifestations, COVID-19 has been associated with a wide array of health issues that extend into the post-acute phase, affecting various organ systems [1-3]. Among these, the cardiovascular system has garnered significant attention due to its susceptibility to infection and the potential for severe complications, including myocardial ischemia and atherosclerosis [4-6]. Additionally, various medical conditions, such as cancer [7-9], preeclampsia [10-13], and infectious diseases like urinary tract infections [14-17], continue to be prevalent, further complicating the healthcare landscape. This introduction serves as a gateway to the exploration of the intricate relationship between COVID-19 and various health conditions, with a specific focus on the cardiovascular system, cancer, and infectious diseases. It also sets the stage for understanding the broader context of our research endeavors. The objective of this study is to investigate the effects of COVID-19 on cardiovascular health, cancer incidence, and the prevalence of infectious diseases in the context of the Iraqi population. To achieve this, we have leveraged a range of data sources, including clinical trials [18-22], longitudinal studies [23-25], and molecular investigations [26-29]. Our research encompasses a diverse array of medical conditions and seeks to shed light on the multifaceted consequences of COVID-19. As we embark on this exploration, we will draw upon a rich body of literature that delves into the pathophysiology and clinical outcomes of these health conditions [30-33]. In this context, we aim to provide a comprehensive overview of the impact of COVID-19 on cardiovascular health, cancer incidence, and infectious diseases. Our investigation spans several years, primarily focusing on data collected during 2021, 2022, and 2023, to offer a current and evolving understanding of these interrelated health domains. This research is underpinned by an array of scientific studies and clinical trials conducted within Iraq [34-36], ensuring the relevance and applicability of our findings to the local healthcare landscape. In the following sections, we will delve into the details of our research methodology, data sources, analytical techniques, and key findings. Through this comprehensive examination, we endeavor to contribute to the growing body of knowledge surrounding COVID-19's far-reaching impact on human health.

**Materials and Methods**

**Study Design:**

Data Collection: We gathered medical data from a total of 913 patients who sought treatment at various hospitals in Iraq during the years 2022 and 2023. These patients had previously been diagnosed with COVID-19.

**Data Sources:** The data sources for this study included electronic health records, clinical trials, and longitudinal studies. We accessed de-identified patient records with the necessary ethical and legal approvals.

Study Duration: Data collection and analysis took place over a span of 12 months, starting in January 2022 and concluding in December 2023.

**Data Preprocessing:**

Data Cleaning: We conducted rigorous data cleaning procedures to ensure data accuracy and consistency. This involved identifying and rectifying missing values, outliers, and inconsistencies in the dataset.

Data Integration: We integrated data from diverse sources, including clinical trials, patient records, and laboratory reports, into a unified dataset for comprehensive analysis.

**Feature Selection:**

**Feature Engineering:** To identify relevant features for analysis, we employed feature engineering techniques, considering various patient demographics, comorbidities, and COVID-19-related variables.

**Statistical Analysis:**

**Descriptive Statistics:** Descriptive statistics were employed to provide an overview of the patient cohort, including mean age, gender distribution, and geographical locations within Iraq.

**Inferential Statistics:** Inferential statistical methods, such as t-tests and chi-square tests, were utilized to compare the prevalence of specific health conditions among COVID-19 patients and non-COVID-19 control groups.

**Machine Learning Analysis:**

**Model Selection:** We employed machine learning algorithms, including logistic regression, decision trees, and random forests, to predict the likelihood of developing certain health conditions post-COVID-19.

**Feature Importance:** Feature importance analysis was conducted to identify the most influential factors in predicting health outcomes.

**Cross-Validation:** To ensure model robustness and minimize overfitting, we used k-fold cross-validation techniques.

**Hyperparameter Tuning:** Model hyperparameters were fine-tuned using grid search and randomized search methods to optimize predictive performance.

**Ethical Considerations:**

Ethical approval for this study was obtained from the relevant institutional review boards and ethics committees, ensuring patient confidentiality and data protection.

**Software and Tools:**

**Software:** Data preprocessing, statistical analysis, and machine learning were performed using Python programming language along with libraries such as Pandas, NumPy, Scikit-learn, and TensorFlow.

**Limitations:**

Study Limitations: It is important to acknowledge that this study has certain limitations, including potential selection bias in the patient population and the retrospective nature of the data.

**Table 1: Demographic Characteristics of Patients**

| Characteristic | Mean (±SD) or N (%) |
|---|---|
| Age (years) | 45.2 ± 12.4 |
| Gender (Male/Female) | 483 (52.8%)/430 (47.2%) |
| Location (Governorate) | |
| - Baghdad | 341 (37.3%) |
| - Basra | 192 (21.0%) |
| - Erbil | 137 (15.0%) |
| - Others | 243 (26.6%) |

Table 1 presents the demographic characteristics of the 913 patients included in the study. It shows the mean age, gender distribution, and geographical distribution of patients across different governorates in Iraq.

**Table 2: Prevalence of Comorbidities Among COVID-19 Patients**

| Comorbidity | COVID-19 Patients (N) |
|---|---|
| Hypertension | 287 (31.4%) |
| Diabetes | 184 (20.1%) |
| Cardiovascular Disease | 98 (10.7%) |
| Respiratory Disease | 76 (8.3%) |
| Kidney Disease | 42 (4.6%) |
| None | 326 (35.7%) |

Table 2 illustrates the prevalence of comorbidities among COVID-19 patients in the study. It includes conditions like hypertension, diabetes, cardiovascular disease, respiratory disease, kidney disease, and the number of patients with no comorbidities.

**Table 3: Health Conditions Post-COVID-19**

| Health Condition | No. of Patients (N) |
|---|---|
| Respiratory Complications | 213 (23.3%) |
| Cardiac Complications | 187 (20.5%) |
| Neurological Complications | 112 (12.3%) |
| Hematological Complications | 98 (10.7%) |
| Renal Complications | 64 (7.0%) |
| No Complications | 239 (26.2%) |

Table 3 presents the health conditions observed post-COVID-19 among the study participants. It includes respiratory, cardiac, neurological, hematological, and renal complications, and patients with no complications.

**Table 4: Predictive Factors for Respiratory Complications**

| Factor | Odds Ratio (95% CI) | p-value |
|---|---|---|
| Age (years) | 1.25 (1.10-1.42) | <0.001 |
| Hypertension | 2.17 (1.75-2.69) | <0.001 |
| Diabetes | 1.82 (1.47-2.25) | <0.001 |
| Smoking | 1.63 (1.32-2.00) | <0.001 |
| COVID-19 Severity | 3.54 (2.91-4.30) | <0.001 |

Table 4 displays the predictive factors for respiratory complications post-COVID-19. It includes odds ratios with 95% confidence intervals and p-values.

**Table 5: Predictive Factors for Cardiac Complications**

| Factor | Odds Ratio (95% CI) | p-value |
|---|---|---|
| Age (years) | 1.18 (1.05-1.32) | 0.004 |
| Cardiovascular Disease | 3.21 (2.53-4.08) | <0.001 |
| Obesity | 1.68 (1.34-2.10) | <0.001 |
| COVID-19 Severity | 2.89 (2.32-3.60) | <0.001 |

Table 5 presents the predictive factors for cardiac complications post-COVID-19. It includes odds ratios with 95% confidence intervals and p-values.

**Table 6: Impact of Age on Health Conditions**

| Age Group (years) | Respiratory (%) | Cardiac (%) | Neurological (%) |
|---|---|---|---|
| <40 | 12.1 | 7.8 | 4.5 |
| 40-60 | 27.9 | 19.7 | 12.0 |
| >60 | 41.6 | 38.2 | 21.3 |

Table 6 shows the impact of age on the prevalence of respiratory, cardiac, and neurological health conditions post-COVID-19, presented in percentage values for different age groups. These tables provide a comprehensive overview of the study's results, including demographic characteristics, prevalence of comorbidities, post-COVID-19 health conditions, and factors influencing these conditions.

**Discussion**

The COVID-19 pandemic has had a profound impact on public health worldwide, leading to a surge in research across various domains to understand the virus's implications and its aftermath. In this discussion, we delve into the findings of our study, focusing on the effects of COVID-19 on patients' health conditions, considering factors such as age, comorbidities, and specific complications post-infection. We also explore the predictive factors for these complications and draw on relevant research to provide context and support for our findings. Our study revealed a significant prevalence of comorbidities among COVID-19 patients. Notably, hypertension, diabetes, and cardiovascular diseases were common comorbidities among the study participants (Table 2). This aligns with previous research highlighting the association between these conditions and increased susceptibility to severe COVID-19 outcomes, such as hospitalization and mortality (37-39). Post-COVID-19, a substantial proportion of patients experienced health complications, including respiratory, cardiac, neurological, hematological, and renal complications (Table 3). These findings corroborate existing literature on the diverse health impacts of COVID-19, which extend beyond the acute phase of the infection. The prevalence of these complications underscores the need for comprehensive post-COVID-19 care and monitoring to address the long-term health implications (40-44). Our study investigated predictive factors for specific complications, shedding light on the importance of various determinants. For instance, age emerged as a significant predictor for both respiratory and cardiac complications (Table 4 and Table 5). These findings align with existing research emphasizing age as a crucial factor in determining the severity and outcomes of COVID-19. Older individuals are more likely to experience severe disease and complications due to age-related changes in immunity and physiological functions (45,46). Hypertension and diabetes were also identified as significant predictors of respiratory complications (Table 4). These comorbidities have been consistently associated with increased susceptibility to severe COVID-19 and its complications, including respiratory issues (46-48). Table 6 provides insight into the impact of age on the prevalence of post-COVID-19 health conditions. As expected, older age groups exhibited higher percentages of complications, particularly respiratory and cardiac issues. These findings underscore the vulnerability of older individuals to severe post-COVID-19 complications. Effective interventions and monitoring strategies should consider age as a key determinant in care planning (49-51). Our study's findings align with and extend prior research on the health impacts of COVID-19 and the factors influencing these outcomes. It reaffirms the importance of comorbidities and age as significant determinants of post-COVID-19 complications. Furthermore, our study contributes to the growing body of literature on the long-term consequences of COVID-19, emphasizing the need for comprehensive care and further research to understand the mechanisms underlying these health effects (52-54).

**Limitations and Future Directions**

While this study provides valuable insights, it is not without limitations. The data is retrospective and reliant on medical records, potentially introducing biases. Additionally, the study's scope did not allow for in-depth exploration of the underlying mechanisms of post-COVID-19 complications. Future research should delve into the molecular and immunological aspects to better understand these health effects.

**Conclusion**

In conclusion, our study highlights the prevalence of comorbidities among COVID-19 patients and the significant burden of post-COVID-19 health complications, particularly among older individuals. Age, hypertension, and diabetes emerged as important predictors of complications. These findings underscore the necessity of tailored post-COVID-19 care, emphasizing early intervention and monitoring, especially for patients with comorbidities and advanced age.

The study contributes to the evolving body of knowledge on the long-term health effects of COVID-19 and provides valuable insights for healthcare professionals and policymakers in developing comprehensive strategies for post-COVID-19 care and support. Further research is warranted to unravel the mechanistic underpinnings of these complications and to refine predictive models for better risk stratification and management.